\documentclass[10pt,twocolumn,letterpaper]{article}

\usepackage{iccv}
\usepackage{times}
\usepackage{epsfig}
\usepackage{graphicx}
\usepackage{amsmath}
\usepackage{amssymb}
\usepackage{algorithm}
\usepackage{algorithmic}
\usepackage{multirow}
\usepackage{subfigure}
\usepackage{enumitem}
\usepackage{colortbl,xcolor}
\DeclareMathOperator*{\argmin}{\arg\min}
\long\def\comment#1{}
\usepackage{lipsum}
\usepackage{makecell}

\usepackage[breaklinks=true,bookmarks=false]{hyperref}

\iccvfinalcopy 

\def\iccvPaperID{3403} 

\ificcvfinal\fi

\begin{document}

\title{MetaPruning: Meta Learning for Automatic Neural Network Channel Pruning}

\author{Zechun Liu$^{1}$\\
\and
Haoyuan Mu$^{2}$\\
\and
Xiangyu Zhang$^{3}$\\
\and
Zichao Guo$^{3}$\\
\and
Xin Yang$^{4}$\\
\and
Tim Kwang-Ting Cheng$^{1}$\\
\and
Jian Sun$^{3}$\\
\and
$^1$ Hong Kong University of Science and Technology 
$^2$ Tsinghua University\\
$^3$ Megvii Technology
$^4$ Huazhong University of Science and Technology}

\maketitle
\ificcvfinal\thispagestyle{empty}\fi
\newcommand\blfootnote[1]{%
\begingroup 
\renewcommand\thefootnote{}\footnote{#1}%
\addtocounter{footnote}{-1}%
\endgroup 
}
\begin{abstract}
In this paper, we propose a novel meta learning approach for automatic channel pruning of very deep neural networks. We first train a PruningNet, a kind of meta network, which is able to generate weight parameters for any pruned structure given the target network. We use a simple stochastic structure sampling method for training the PruningNet. Then, we apply an evolutionary procedure to search for good-performing pruned networks. The search is highly efficient because the weights are directly generated by the trained PruningNet and we do not need any finetuning at search time. With a single PruningNet trained for the target network, we can search for various Pruned Networks under different constraints with little human participation. Compared to the state-of-the-art pruning methods, we have demonstrated superior performances on MobileNet V1/V2 and ResNet. Codes are available on \url{https://github.com/liuzechun/MetaPruning}.
\blfootnote{This work is done when Zechun Liu and Haoyuan Mu are interns at Megvii Technology.}
\end{abstract}

\section{Introduction}
Channel pruning has been recognized as an effective neural network compression/acceleration method~\cite{pruning_filters, channel_pruning, structured_pruning, coarse_pruning, amc, netadapt} and is widely used in the industry.
A typical pruning approach contains three stages: training a large over-parameterized network, pruning the less-important weights or channels, finetuning or re-training the pruned network. The second stage is the key. It usually performs iterative layer-wise pruning and fast finetuning or weight reconstruction to retain the accuracy~\cite{learning_both_weights_and_connections, learning_number_of_neuron, sparse_cnn, pruning_CNN}. 

Conventional channel pruning methods mainly rely on data-driven sparsity constraints~\cite{data-driven, network_slimming}, or human-designed policies~\cite{channel_pruning, pruning_filters, diversity_networks, network_trimming, thinet, structured_pruning}.
Recent AutoML-style works automatically prune channels in an iterative mode, based on a feedback loop~\cite{netadapt} or reinforcement learning~\cite{amc}. Compared with the conventional pruning methods, the AutoML methods save human efforts and can optimize the direct metrics like the hardware latency. 

Apart from the idea of keeping the important weights in the pruned network, a recent study~\cite{rethink_pruning} finds that the pruned network can achieve the same accuracy no matter it inherits the weights in the original network or not. This finding suggests that the essence of channel pruning is finding good pruning structure - layer-wise channel numbers.

\begin{figure}[t]
\centering
\includegraphics[width=\linewidth]{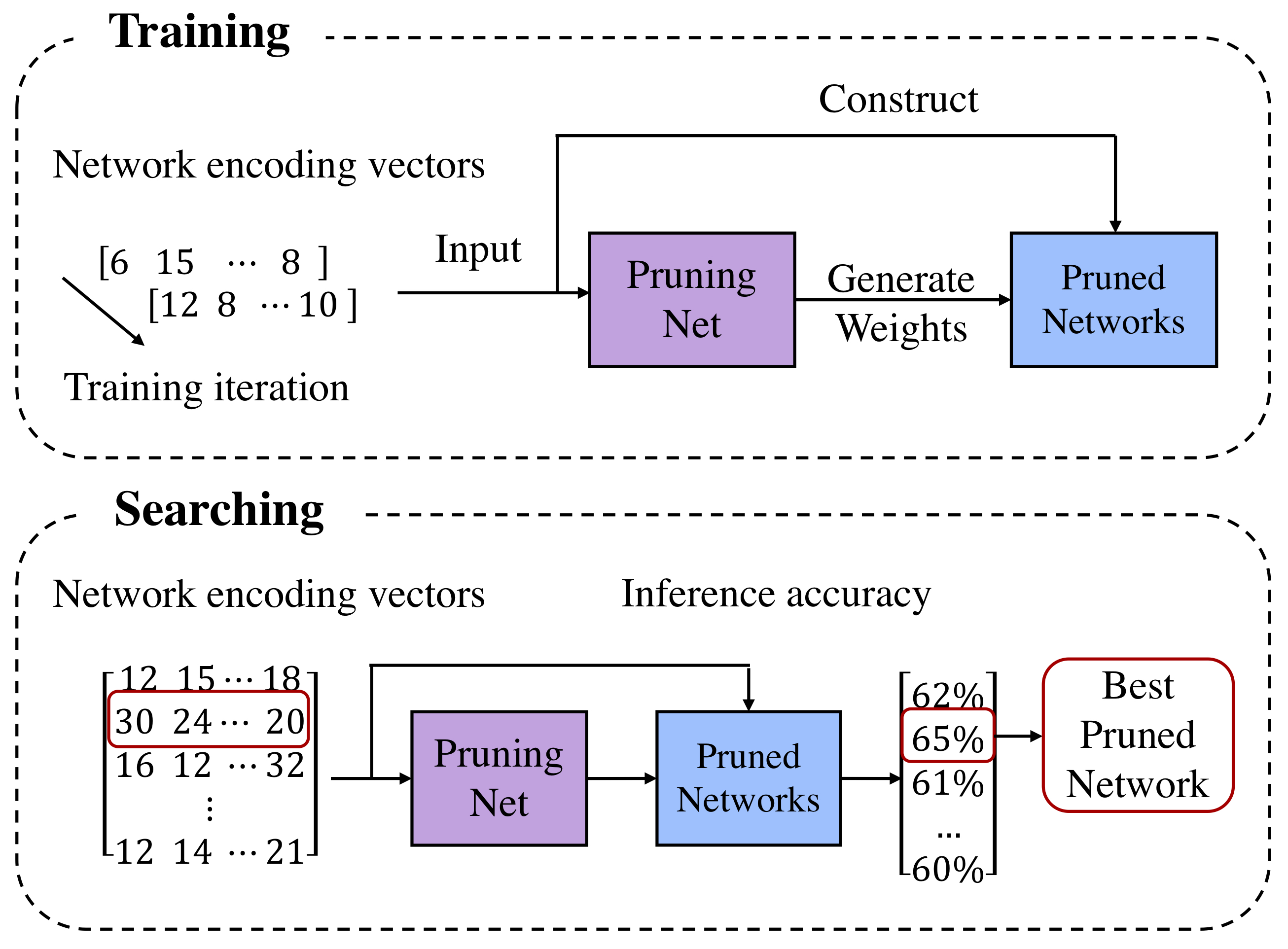}
\caption{Our MetaPruning has two steps. 1) training a PruningNet. At each iteration, a network encoding vector (i.e., the number of channels in each layer) is randomly generated. The Pruned Network is constructed accordingly. The PruningNet takes the network encoding vector as input and generates the weights for the Pruned Network. 2) searching for the best Pruned Network. We construct many Pruned Networks by varying network encoding vector and evaluate their goodness on the validation data with the weights predicted by the PruningNet. No finetuning or re-training is needed at search time.}
\vspace{-1em}
\label{fig:high_level}
\end{figure}

However, exhaustively finding the optimal pruning structure is computationally prohibitive. Considering a network with 10 layers and each layer contains 32 channels. The possible combination of layer-wise channel numbers could be $32^{10}$. Inspired by the recent Neural Architecture Search (NAS), specifically One-Shot model~\cite{one_shot}, as well as the weight prediction mechanism in HyperNetwork~\cite{hypernetworks}, we propose to train a \emph{PruningNet} that can generate weights for all candidate pruned networks structures, such that we can search good-performing structures by just evaluating their accuracy on the validation data, which is highly efficient. 

To train the PruningNet, we use a stochastic structure sampling.
As shown in Figure \ref{fig:high_level}, the PruningNet generates the weights for pruned networks with corresponding network encoding vectors, which is the number of channels in each layer. By stochastically feeding in different network encoding vectors, the PruningNet gradually learns to generate weights for various pruned structures.
After the training, we search for good-performing Pruned Networks by an evolutionary search method which can flexibly incorporate various constraints such as computation FLOPs or hardware latency. Moreover, by directly searching the best pruned network via determining the channels for each layer or each stage, we can prune channels in the shortcut without extra effort, which is seldom addressed in previous channel pruning solutions. We name the proposed method as MetaPruning.

We apply our approach on MobileNets~\cite{mobilenet_v1, mobilenet_v2} and ResNet~\cite{resnet}. At the same FLOPs, our accuracy is $2.2\%$-$6.6\%$ higher than MobileNet V1, $0.7\%$-$3.7\%$ higher than MobileNet V2, and $0.6\%$-$1.4\%$ higher than ResNet-50. At the same latency, our accuracy is $2.1\%$-$9.0\%$ higher than MobileNet V1, and $1.2\%$-$9.9\%$ higher than MobileNet V2. Compared with state-of-the-art channel pruning methods~\cite{amc,netadapt}, our MetaPruning also produces superior results.

Our contribution lies in four folds:
\begin{itemize}[leftmargin=*]\setlength{\itemsep}{-2pt}
\item We proposed a meta learning approach, MetaPruning, for channel pruning. The central of this approach is learning a meta network (named PruningNet) which generates weights for various pruned structures. With a single trained PruningNet, we can search for various pruned networks under different constraints.
\item Compared to conventional pruning methods, MetaPruning liberates human from cumbersome hyperparameter tuning and enables the direct optimization with desired metrics.
\item Compared to other AutoML methods, MetaPruning can easily enforce constraints in the search of desired structures, without manually tuning the reinforcement learning hyper-parameters.
\item The meta learning is able to effortlessly prune the channels in the short-cuts for ResNet-like structures, which is non-trivial because the channels in the short-cut affect more than one layers.
\end{itemize}
\section{Related Works}
There are extensive studies on compressing and accelerating neural networks, such as quantization~\cite{lq-net, xnornet, birealnet, loss-aware-quantization, elq, zhuang2019structured}, pruning~\cite{channel_pruning, brain_damage, deep_compression} and compact network design~\cite{mobilenet_v1, mobilenet_v2, shufflenet_v1, shufflenet_v2, squeezenet}. A comprehensive survey is provided in~\cite{efficient_dnn_survey}. Here, we summarize the approaches that are most related to our work.

\textbf{Pruning}
Network pruning is a prevalent approach for removing redundancy in DNNs. In weight pruning, people prune individual weights to compress the model size~\cite{brain_damage, brain_surgeon, deep_compression, dynamic_surgery}. However, weight pruning results in unstructured sparse filters, which can hardly be accelerated by general-purpose hardware. Recent works~\cite{network_trimming, pruning_filters, diversity_networks, channel_pruning, thinet, slimmable} focus on channel pruning in the CNNs, which removes entire weight filters instead of individual weights. Traditional channel pruning methods trim channels based on the importance of each channel either in an iterative mode~\cite{channel_pruning, thinet} or by adding a data-driven sparsity~\cite{data-driven,network_slimming}. In most traditional channel pruning, compression ratio for each layer need to be manually set based on human experts or heuristics, which is time consuming and prone to be trapped in sub-optimal solutions.

\textbf{AutoML}
Recently, AutoML methods~\cite{amc, netadapt, chen2018constraint, dong2018dpp} take the real-time inference latency on multiple devices into account to iteratively prune channels in different layers of a network via reinforcement learning~\cite{amc} or an automatic feedback loop~\cite{netadapt}. Compared with traditional channel pruning methods, AutoML methods help to alleviate the manual efforts for tuning the hyper-parameters in channel pruning.
Our proposed MetaPruning also involves little human participation. Different from previous AutoML pruning methods, which is carried out in a layer-wise pruning and finetuning loop, our methods is motivated by recent findings~\cite{rethink_pruning}, which suggests that instead of selecting ``important'' weights, the essence of channel pruning sometimes lies in identifying the best pruned network. 
From this prospective, we propose MetaPruning for directly finding the optimal pruned network structures. Compared to previous AutoML pruning methods~\cite{amc, netadapt}, MetaPruning method enjoys higher flexibility in precisely meeting the constraints and possesses the ability of pruning the channel in the short-cut. 

\textbf{Meta Learning}
Meta-learning refers to learning from observing how different machine learning approaches perform on various learning tasks. Meta learning can be used in few/zero-shot learning~\cite{fewshot, zeroshot} and transfer learning~\cite{learning_to_learn}. A comprehensive overview of meta learning is provided in~\cite{metalearning_survey}.
In this work we are inspired by~\cite{hypernetworks} to use meta learning for weight prediction. Weight predictions refer to weights of a neural network are predicted by another neural network rather than directly learned~\cite{hypernetworks}. Recent works also applies meta learning on various tasks and achieves state-of-the-art results in detection~\cite{metaanchor}, super-resolution with arbitrary magnification~\cite{metasr} and instance segmentation~\cite{learning_to_seg}. 

\textbf{Neural Architecture Search}
Studies for neural architecture search try to find the optimal network structures and hyper-parameters with reinforcement learning~\cite{NAS_with_RL, Design_NN_using_RL}, genetic algorithms~\cite{genetic_CNN, ENAS, large_scale_evolution} or gradient based approaches~\cite{darts, fbnet}. 
Parameter sharing~\cite{proxylessnas, one_shot, fbnet, darts} and weights prediction~\cite{smash, predicting_parameters} methods are also extensively studied in neural architecture search. One-shot architecture search~\cite{one_shot} uses an over-parameterized network with multiple operation choices in each layer. By jointly training multiple choices with drop-path, it can search for the path with highest accuracy in the trained network, which also inspired our two step pruning pipeline.
Tuning channel width are also included in some neural architecture search methods.
ChamNet~\cite{chamnet} built an accuracy predictor atop Gaussian Process with Bayesian optimization to predict the network accuracy with various channel widths, expand ratios and numbers of blocks in each stage. Despite its high accuracy, building such an accuracy predictor requires a substantial of computational power.
FBNet~\cite{fbnet} and ProxylessNas~\cite{proxylessnas} include blocks with several different middle channel choices in the search space. Different from neural architecture search, in channel pruning task, the channel width choices in each layer is consecutive, which makes enumerate every channel width choice as an independent operation infeasible. Proposed MetaPruning targeting at channel pruning is able to solve this consecutive channel pruning challenge by training the PruningNet with weight prediction, which will be explained in Sec.\ref{sec:methodology} 

\begin{figure}[t]
\centering
\includegraphics[width=0.9\linewidth]{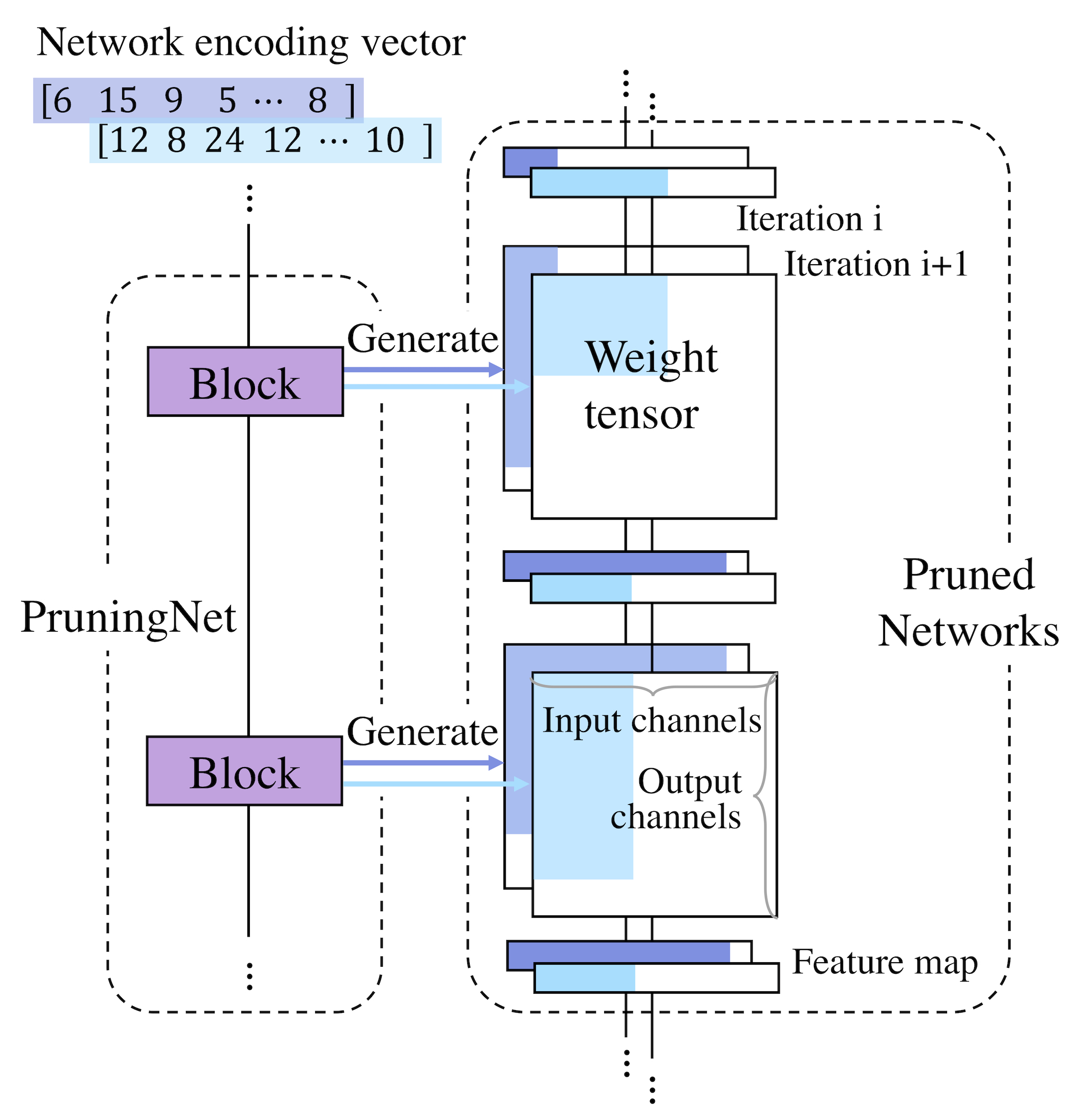}
\caption{The proposed stochastic training method of PruningNet. At each iteration, we randomize a network encoding vector. The PruningNet generates the weight by taking the vector as input. The Pruned Network is constructed with respect to the vector. We crop the weights generated by the PruningNet to match the input and output channels in the Pruned Networks. By change network encoding vector in each iteration, the PruningNet can learn to generate different weights for various Pruned Networks.}
\vspace{-2em}
\label{fig:overall}
\end{figure}

\section{Methodology}
\label{sec:methodology}
In this section, we introduce our meta learning approach for automatically pruning channels in deep neural networks, that pruned network could meet various constraints easily.

We formulate the channel pruning problem as 
\begin{equation}
\begin{aligned}
& (c_1, c_2, ... c_l)^* = \argmin_{c_1, c_2, ... c_l} \mathcal{L}(\mathcal{A}(c_1, c_2, ... c_l; w)) \\
& s.t. ~~~\mathcal{C} < constraint,
\label{pruning_problem}
\end{aligned}
\end{equation}
where $\mathcal{A}$ is the network before the pruning. We try to find out the pruned network channel width ($c_1, c_2,  ..., c_l$) for $1^{st}$ layer to $l^{th}$ layer that has the minimum loss after the weights are trained, with the cost $\mathcal{C}$ meets the constraint (i.e. FLOPs or latency).

To achieve this, we propose to construct a PruningNet, a kind of meta network, where we can quickly obtain the goodness of all potential pruned network structures by evaluating on the validation data only. Then we can apply any search method, which is evolution algorithm in this paper, to search for the best pruned network.

\subsection{PruningNet training} 

Channel pruning is non-trivial because the layer-wise dependence in channels such that pruning one channel may significantly influence the following layers and, in return, degrade the overall accuracy. Previous methods try to decompose the channel pruning problem into the sub-problem of pruning the unimportant channels layer-by-layer~\cite{channel_pruning} or adding the sparsity regularization~\cite{data-driven}. AutoML methods prune channels automatically with a feedback loop~\cite{netadapt} or reinforcement learning~\cite{amc}. Among those methods, how to prune channels in the short-cut is seldom addressed. Most previous methods prune the middle channels in each block only\cite{netadapt, amc}, which limits the overall compression ratio. 

Carrying out channel pruning task with consideration of the overall pruned network structure is beneficial for finding optimal solutions for channel pruning and can solve the shortcut pruning problem. However, obtaining the best pruned network is not straightforward, considering a small network with 10 layers and each layer containing 32 channels, the combination of possible pruned network structures is huge.

Inspired by the recent work~\cite{rethink_pruning}, which suggests the weights left by pruning is not important compared to the pruned network structure, we are motivated to directly find the best pruned network structure. In this sense, we may directly predict the optimal pruned network without iteratively decide the important weight filters. To achieve this goal, we construct a meta network, PruningNet, for providing reasonable weights for various pruned network structures to rank their performance.

The PruningNet is a meta network, which takes a network encoding vector $(c_1, c_2, ... c_l)$ as input and outputs the weights of pruned network: 
\begin{equation}
\begin{aligned}
\mathbf{W} = \emph{PruningNet}(c_1, c_2, ... c_l).
\label{eq:dicionary_network}
\end{aligned}
\end{equation}

\begin{figure}[t]
\centering
\includegraphics[width=\linewidth]{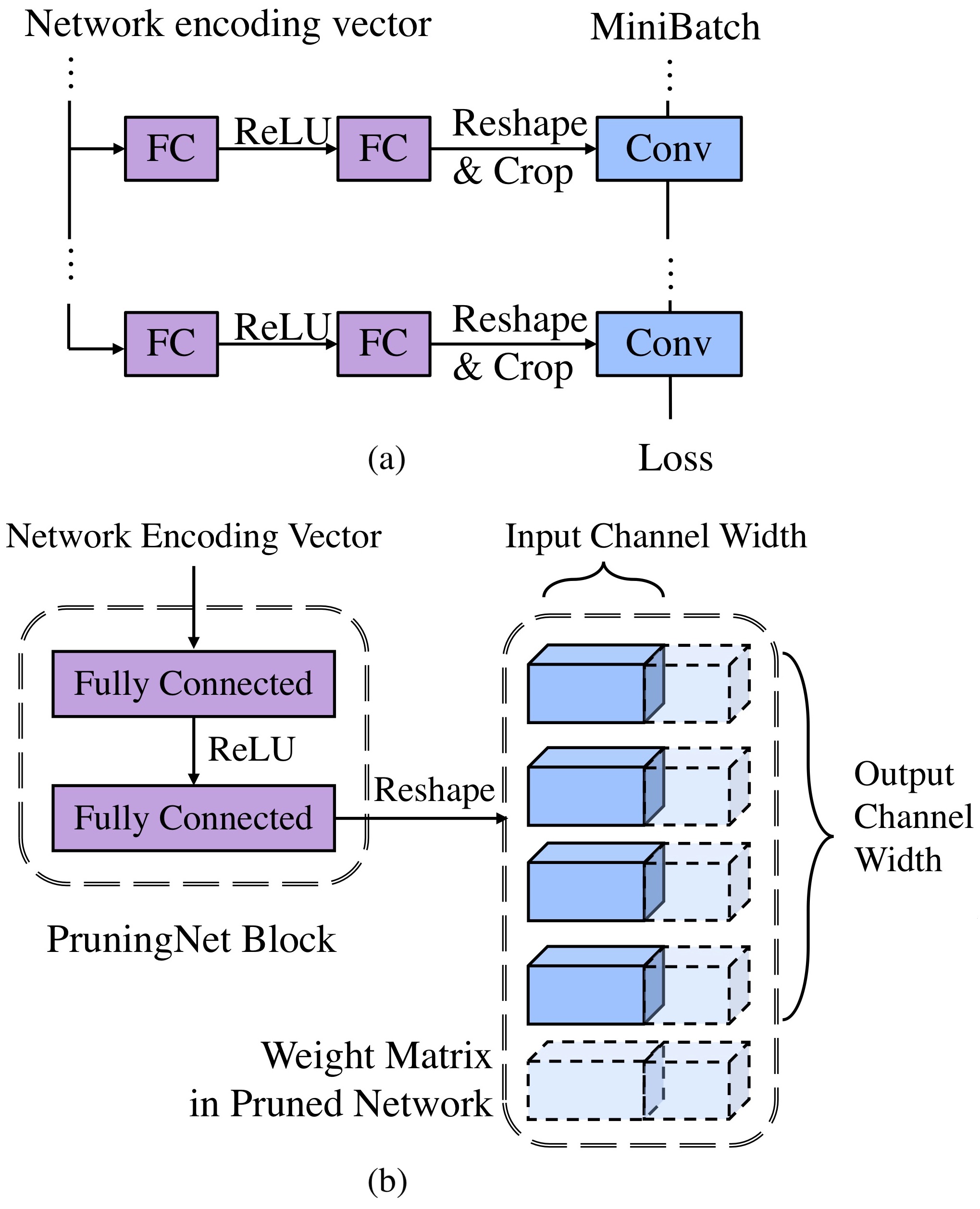}
\caption{(a) The network structure of PruningNet connected with Pruned Network. The PruningNet and the Pruned Network are jointly trained with input of the network encoding vector as well as a mini-batch of images. (b) The reshape and crop operation on the weight matrix generated by the PruningNet block.}
\vspace{-1em}
\label{fig:meta}
\end{figure}

A PruningNet block consists of two fully-connected layers.
In the forward pass, the PruningNet takes the network encoding vector (i.e., the number of channels in each layer) as input, and generates the weight matrix. Meanwhile, a Pruned Network is constructed with output channels width in each layer equal to the element in the network encoding vector. The generated weight matrix is cropped to match the number of input and output channel in the Pruned Network, as shown in Figure \ref{fig:overall}. Given a batch of input image, we can calculate the loss from the Pruned Network with generated weights.

In the backward pass, instead of updating the weights in the Pruned Networks, we calculate the gradients w.r.t the weights in the PruningNet. 
Since the reshape operation as well as the convolution operation between the output of the fully-connect layer in the PruningNet and the output of the previous convolutional layer in the Pruned Network is also differentiable, the gradient of the weights in the PruningNet can be easily calculated by the Chain Rule. The PruningNet is end-to-end trainable. The detailed structure of PruningNet connected with Pruned Network is shown in Figure \ref{fig:meta}.

To train the PruningNet, we proposed the stochastic structure sampling. In the training phase, the network encoding vector is generated by randomly choosing the number of channels in each layer at each iteration. With different network encodings, different Pruned Networks are constructed and the corresponding weights are provided with the PruningNet. By stochastically training with different encoding vectors, the PruningNet learns to predict reasonable weights for various different Pruned Networks.

\subsection{Pruned-Network search}
After the PruningNet is trained, we can obtain the accuracy of each potential pruned network by inputting the network encoding into the PruningNet, generating the corresponding weights and doing the evaluation on the validation data.

Since the number of network encoding vectors is huge, we are not able to enumerate. To find out the pruned network with high accuracy under the constraint, we use an evolutionary search, which is able to easily incorporate any soft or hard constraints. 

In the evolutionary algorithm used in MetaPruning, each pruned network is encoded with a vector of channel numbers in each layer, named the genes of pruned networks. Under the hard constraint, we first randomly select a number of genes and obtain the accuracy of the corresponding pruned network by doing the evaluation. Then the top k genes with highest accuracy are selected for generating the new genes with mutation and crossover. The mutation is carried out by changing a proportion of elements in the gene randomly. The crossover means that we randomly recombine the genes in two parent genes to generate an off-spring. We can easily enforce the constraint by eliminate the unqualified genes. By further repeating the top k selection process and new genes generation process for several iterations, we can obtain the gene that meets constraints while achieving the highest accuracy. Detailed algorithm is described in Algorithm.\ref{alg:1}.

\begin{algorithm}[ht]
\caption{Evolutionary Search Algorithm}
\label{alg:1}
\textbf{Hyper Parameters}: Population Size: $\mathcal{P}$, Number of Mutation: $\mathcal{M}$, Number of Crossover: $\mathcal{S}$, Max Number of Iterations: $\mathcal{N}$ . \\
\textbf{Input}: PruningNet: $PruningNet$, Constraints: $\mathcal{C}$ .\\
\textbf{Output}: Most accurate gene: $\mathcal{G}_{top}$ .\\
\begin{algorithmic}[1]
\STATE $\mathcal{G}_0$ = Random($\mathcal{P}$), s.t. $\mathcal{C}$; 
\STATE $\mathcal{G}_{topK}$ = $\emptyset$; 
\FOR{$i = 0:\mathcal{N}$}
\STATE \{$\mathcal{G}_i$, accuracy\} = Inference($PruningNet(\mathcal{G}_i$));\
\STATE $\mathcal{G}_{topK}$, $accuracy_{topK}$ = TopK(\{$\mathcal{G}_i$, $accuracy$\});\
\STATE $\mathcal{G}_{mutation}$ = Mutation($\mathcal{G}_{topK}, \mathcal{M}$), s.t. $\mathcal{C}$;\
\STATE $\mathcal{G}_{crossover}$ = Crossover($\mathcal{G}_{topK},  \mathcal{S}$), s.t. $\mathcal{C}$;\
\STATE $\mathcal{G}_{i}$ = $\mathcal{G}_{mutation}$ + $\mathcal{G}_{crossover}$;\
\ENDFOR
\STATE $\mathcal{G}_{top1}$, $accuracy_{top1}$= Top1(\{$\mathcal{G}_{\mathcal{N}}$, $accuracy$\});\
\STATE \textbf{return} $\mathcal{G}_{top1}$;
\end{algorithmic}
\end{algorithm}

\section{Experimental Results}
In this section, we demonstrate the effectiveness of our proposed MetaPruning method. 
We first explain the experiment settings and introduce how to apply the MetaPruning on MobileNet V1~\cite{mobilenet_v1} V2~\cite{mobilenet_v2} and ResNet~\cite{resnet}, which can be easily generalized to other network structures. Second, we compare our results with the uniform pruning baselines as well as state-of-the-art channel pruning methods. Third, we visualize the pruned network obtained with MetaPruning. Last, ablation studies are carried out to elaborate the effect of weight prediction in our method.

\subsection{Experiment settings}
\label{sec:experiment_settings}
The proposed MetaPruning is very efficient. Thus it is feasible to carry out all experiments on the ImageNet 2012 classification dataset~\cite{imagenet}. 

MetaPruning method consists of two stages. In the first stage, the PruningNet is train from scratch with stochastic structure sampling, which takes $\frac{1}{4}$ epochs as training a network normally. Further prolonging PruningNet training yields little final accuracy gain in the obtained Pruned Net. In the second stage, we use an evolutionary search algorithm to find the best pruned network. With the PruningNet predicting the weights for all the PrunedNets, no fine-tuning or retraining are needed at search time, which makes the evolution search highly efficient. Inferring a PrunedNet only takes seconds on 8 Nvidia 1080Ti GPUs. The best PrunedNet obtained from search is then trained from scratch. For the training process in both stages, we use the standard data augmentation strategies as~\cite{resnet} to process the input images. We adopt the same training scheme as~\cite{shufflenet_v2} for experiments on MobileNets and the training scheme in~\cite{resnet} for ResNet. The resolutions of the input image is set to 224 $\times$ 224 for all experiments. 

At training time, we split the original training images into sub-validation dataset, which contains 50000 images randomly selected from the training images with 50 images in each 1000-class, and sub-training dataset with the rest of images. We train the PruningNet on the sub-training dataset and evaluating the performance of pruned network on the sub-validation dataset in the searching phase. At search time, we recalculate the running mean and running variance in the BatchNorm layer with 20000 sub-training images for correctly inferring the performance of pruned networks, which takes only a few seconds. After obtaining the best pruned network, the pruned network is trained from scratch on the original training dataset and evaluated on the test dataset.

\subsection{MetaPruning on MobileNets and ResNet}
To prove the effectiveness of our MetaPruning method, we apply it on MobileNets~\cite{mobilenet_v1,mobilenet_v2} and ResNet~\cite{resnet}. 

\begin{figure}[t]
\centering
\includegraphics[width=1\linewidth]{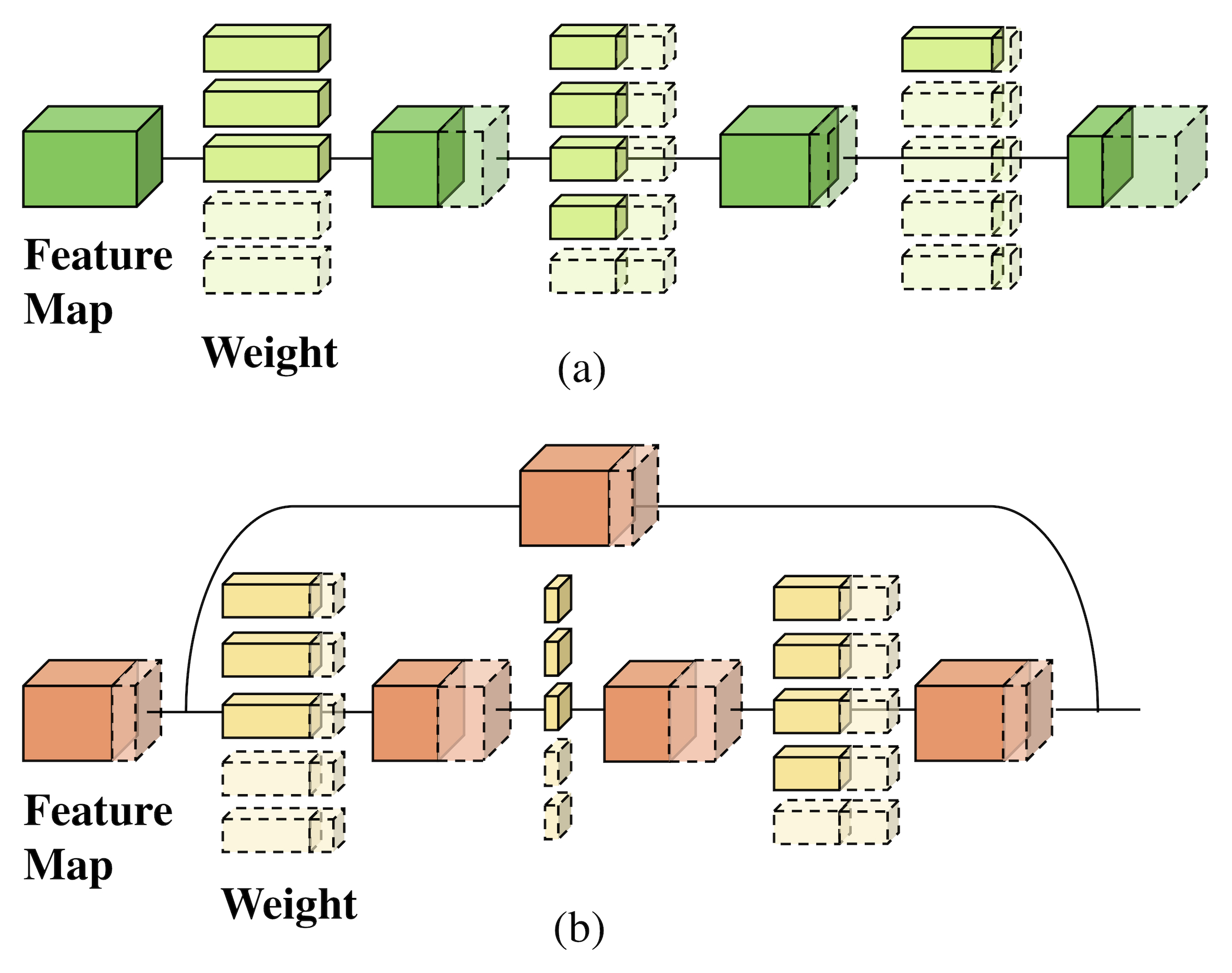}
\caption{Channel Pruning schemes considering the layer-wise inter-dependency. (a) For the network without shortcut, \textit{e.g.}, MobileNet V1, we crop the top left of the original weight matrix to match the input and output channels. For simplification, we omit the depth-wise convolution here; (b) For the network with shortcut, \textit{e.g.}, MobileNet V2, ResNet, we prune the middle channels in the blocks while keep the input and output of the block being equal.}
\vspace{-1em}
\label{fig:channel_pruning_scheme}
\end{figure}

\subsubsection{MobileNet V1}

MobileNet V1 is a network without shortcut. 
To construct the corresponding PruningNet, we have the PruningNet blocks equal to the number of convolution layers in the MobileNet v1, and each PruningNet block is composed of two concatenated fully-connected(FC) layers. 

The input vector to the PruningNet is the number of channels in each layer. Then this vector is decoded into the input and output channel compression ratio of each layer, i.e., $[\frac{C^{l-1}_{po}}{C^{l-1}_{o}}, \frac{C^l_{po}}{C^l_{o}}]$. Here, $C$ denotes the number of channels, $l$ is layer index of current layer and $l-1$ denotes the previous layer, $o$ means output of the original network and $po$ is the pruned output. This two dimensional vector is then inputted into each PruningNet block associated with each layer. The first FC layer in the PruningNet block output a vector with 64 entries and the second FC layer use this 64-entry encoding to output a vector with a length of $C_{o}^l \times C_{o}^{l-1} \times W^l \times H^l$. Then we reshape it to ($C_{o}^l, C_{o}^{l-1}, W^l, H^l$) as the weight matrix in the convolution layer, as shown in Figure.\ref{fig:meta}.

In stochastic structure sampling, an encoding vector of output channel numbers is generated with its each entry $C_{po}^l$ independently and randomly selected from [$int(0.1\times C^l_{o})$, $C^l_{o}$], with the step being $int(0.03\times C^l_{o})$. More refined or coarse step can be chosen according to the fineness of pruning. After decoding and the weight generation process in the PruningNet, the top left part of generated weight matrix is cropped to ($C_{po}^l, C_{po}^{l-1}, W^l, H^l$) and is used in training, and the rest of the weights can be regards as being `untouched' in this iteration, as shown in Figure.\ref{fig:channel_pruning_scheme} (a). In different iterations, different channel width encoding vectors are generated.

\subsubsection{MobileNet V2}
In MobileNet V2, each stage starts with a bottleneck block matching the dimension between two stages. If a stage consists of more than one block, the following blocks in this stage will contain a shortcut adding the input feature maps with the output feature maps, thus input and output channels in a stage should be identical, as shown in Figure \ref{fig:channel_pruning_scheme} (b). To prune the structure containing shortcut, we generate two network encoding vectors, one encodes the overall stage output channels for matching the channels in the shortcut and another encodes the middle channels of each blocks.
In PruningNet, we first decode this network encoding vector to the input, output and middle channel compression ratio of each block. Then we generate the corresponding weight matrices in that block, with a vector $[\frac{C^{b-1}_{po}}{C^{b-1}_{p}}, \frac{C^b_{po}}{C^b_{o}}, \frac{C^b_{middle \ po}}{C^b_{middle \ o}}]$ inputting to the corresponding PruningNet blocks, where $b$ denotes the block index.
The PruningNet block design is the same as that in MobileNetV1, and the number of PruningNet block equals to the number of convolution layers in the MobileNet v2. 

\subsubsection{ResNet}
As a network with shortcut, ResNet has similar network structure with MobileNet v2 and only differs at the type of convolution in the middle layer, the downsampling block and number of blocks in each stage. Thus, we adopt similar PruningNet design for ResNet as MobileNet V2.

\subsection{Comparisons with state-of-the-arts}
We compare our method with the uniform pruning baselines, traditional pruning methods as well as state-of-the-art channel pruning methods.

\comment{- \textbf{Uniform Baselines}
MobileNets~\cite{mobilenet_v1, mobilenet_v2} proposes to use multipliers to uniformly prune the channel width in each layer to meet the resource constraints.

- \textbf{AMC}
\cite{amc} utilizes reinforcement learning to iteratively prune channels in each layer taking the accuracy as well as the latency as the reward.

- \textbf{NetAdapt}
\cite{netadapt} automatically decide pruning how much proportion in which layer based on a feedback loop with latency estimated from the device. }

\subsubsection{Pruning under FLOPs constraint}

\setlength{\tabcolsep}{1pt}
\begin{table}
\begin{center}
\caption{This table compares the top-1 accuracy of MetaPruning method with the uniform baselines on MobileNet V1~\cite{mobilenet_v1}.}
\label{table:mbv1_FLOPs}
\begin{tabular}{ccccccccc}
\noalign{\smallskip}
\hline
\multicolumn{3}{c}{\textbf{Uniform Baselines}} & \multicolumn{2}{c}{\textbf{MetaPruning}} \\
\hline
Ratio \ \ \ & Top1-Acc \ \ \ & FLOPs \ \ \ & Top1-Acc \ \ \ & FLOPs \ \ \ \\
\hline
\hline
1$\times$ & 70.6\% & 569M & -- & --  \\
\hline
0.75$\times$ & 68.4\% & 325M & \textbf{70.9\%} & 324M \\
\hline
0.5$\times$ & 63.7\% & 149M & \textbf{66.1\%} & 149M \\
\hline
0.25$\times$ & 50.6\% & 41M  & \textbf{57.2\%} & 41M \\
\hline
\end{tabular}
\end{center}
\end{table}
\setlength{\tabcolsep}{1.4pt}

\setlength{\tabcolsep}{1pt}
\begin{table}
\begin{center}
\caption{This table compares the top-1 accuracy of MetaPruning method with the uniform baselines on MobileNet V2~\cite{mobilenet_v2}. MobileNet V2 only reports the accuracy with 585M and 300M FLOPs, so we apply the uniform pruning method on MobileNet V2 to obtain the baseline accuracy for networks with other FLOPs.}
\label{table:mbv2_FLOPs}
\begin{tabular}{ccccccccc}
\noalign{\smallskip}
\hline
\multicolumn{2}{c}{\textbf{Uniform Baselines}} & \multicolumn{2}{c}{\textbf{MetaPruning}} \\
\hline
 Top1-Acc \ \ \ & FLOPs \ \ \ & Top1-Acc \ \ \ & FLOPs \ \ \ \\
\hline
\hline
74.7\% & 585M & -- & --  \\
\hline
72.0\% & 313M & \textbf{72.7\%} & 291M \\
\hline
67.2\% & 140M & \textbf{68.2\%} & 140M \\
\hline
66.5\% & 127M & \textbf{67.3\%} & 124M \\
\hline
64.0\% & 106M & \textbf{65.0\%} & 105M \\
\hline
62.1\% & 87M & \textbf{63.8\%} & 84M \\
\hline
54.6\% & 43M & \textbf{58.3\%} & 43M \\
\hline
\end{tabular}
\end{center}
\end{table}
\setlength{\tabcolsep}{1.4pt}

\setlength{\tabcolsep}{1pt}
\begin{table}
\begin{center}
\caption{This table compares the Top-1 accuracy of MetaPruning, uniform baselines and state-of-the-art channel pruning methods, ThiNet~\cite{thinet}, CP~\cite{channel_pruning} and SFP~\cite{he2018soft} on ResNet-50~\cite{resnet}}
\label{table:resnet}
\begin{tabular}{ccccccccc}
\noalign{\smallskip}
\hline
\multicolumn{2}{c}{\textbf{Network}} & \textbf{\ FLOPs\ } & \textbf{\ Top1-Acc\ } \\
\hline
\hline
\multirowcell{3}{Uniform \\ Baseline} {} & \multicolumn{1}{c}{1.0$\times$ ResNet-50\ \ } & 4.1G & 76.6\%  \\
 & \multicolumn{1}{c}{0.75$\times$ ResNet-50} & 2.3G & $74.8\%$ \\
 & \multicolumn{1}{c}{0.5 $\times$ ResNet-50} & 1.1G & $72.0\%$ \\
\hline
\multirowcell{5}{Traditional\\ Pruning}  & SFP\cite{he2018soft} & 2.9G & $75.1\%$ \\
& ThiNet-70~\cite{thinet} & 2.9G & $75.8\%$ \\
& ThiNet-50~\cite{thinet} & 2.1G & $74.7\%$ \\
& ThiNet-30~\cite{thinet} & 1.2G & $72.1\%$ \\
& CP~\cite{channel_pruning} & 2.0G & $73.3\%$\\
\hline
\multicolumn{2}{l}{MetaPruning - 0.85$\times$ResNet-50}  & 3.0G  & $\mathbf{76.2\%}$ \\
\multicolumn{2}{l}{MetaPruning - 0.75$\times$ResNet-50}  & 2.0G  & $\mathbf{75.4\%}$ \\
\multicolumn{2}{l}{MetaPruning - 0.5 $\times$ResNet-50}  & 1.0G  & $\mathbf{73.4\%}$ \\
\hline
\end{tabular}
\end{center}
\end{table}
\setlength{\tabcolsep}{1.4pt}

\setlength{\tabcolsep}{1pt}
\begin{table}
\begin{center}
\caption{This table compares the top-1 accuracy of MetaPruning method with other state-of-the-art AutoML-based methods.}
\label{table:SOA_FLOPs}
\begin{tabular}{ccccccccc}
\noalign{\smallskip}
\hline
\textbf{Network} \ \ \ & \textbf{FLOPs} \ \ \ & \textbf{Top1-Acc} \ \ \ \\
\hline
\hline
0.75x MobileNet V1~\cite{mobilenet_v1} & 325M & 68.4\% \\
NetAdapt~\cite{netadapt} & 284M & 69.1\% \\
AMC~\cite{amc} & 285M & 70.5\% \\
MetaPruning  & 281M & \textbf{70.6}\%  \\
\hline
0.75x MobileNet V2~\cite{mobilenet_v2} & 220M & 69.8\% \\
AMC~\cite{amc} & 220M & 70.8\% \\
MetaPruning  & 217M & \textbf{71.2\%} \\
\hline
\end{tabular}
\end{center}
\end{table}
\setlength{\tabcolsep}{1.4pt}

Table \ref{table:mbv1_FLOPs} compares our accuracy with the uniform pruning baselines reported in \cite{mobilenet_v1}. With the pruning scheme learned by MetaPruning, we obtain 6.6\% higher accuracy than the baseline 0.25$\times$ MobileNet V1. Further more, as our method can be generalized to prune the shortcuts in a network, we also achieves decent improvement on MobileNet V2, shown in Table.\ref{table:mbv2_FLOPs}
Previous pruning methods only prunes the middle channels of the bottleneck structure~\cite{netadapt, amc}, which limits their maximum compress ratio at given input resolution. With MetaPruning, we can obtain 3.7\% accuracy boost when the model size is as small as 43M FLOPs. 
For heavy models as ResNet, MetaPruning also outperforms the uniform baselines and other traditional pruning methods by a large margin, as is shown in Table.\ref{table:resnet}.

In Table \ref{table:SOA_FLOPs}, we compare MetaPruning with the state-of-the-art AutoML pruning methods. MetaPruning achieves superior results than AMC~\cite{amc} and NetAdapt~\cite{netadapt}. Moreover, MetaPruning gets rid of manually tuning the reinforcement learning hyper-parameters and can obtain the pruned network precisely meeting the FLOPs constraints. With the PruningNet trained once using one-fourth epochs as normally training the target network, we can obtain multiple pruned network structures to strike different accuracy-speed trade-off, which is more efficient than the state-of-the-art AutoML pruning methods~\cite{amc, netadapt}. The time cost is reported in Sec.\ref{sec:experiment_settings}.

\subsubsection{Pruning under latency constraint} 

There is an increasing attention in directly optimizing the latency on the target devices. Without knowing the implementation details inside the device, MetaPruning learns to prune channels according to the latency estimated from the device. 

As the number of potential Pruned Network is numerous, measuring the latency for each network is too time-consuming. With a reasonable assumption that the execution time of each layer is independent, we can obtain the network latency by summing up the run-time of all layers in the network. Following the practice in~\cite{fbnet, netadapt}, we first construct a look-up table, by estimating the latency of executing different convolution layers with different input and output channel width on the target device, which is Titan Xp GPU in our experiments. Then we can calculate the latency of the constructed network from the look-up table.

We carried out experiments on MobileNet V1 and V2. Table \ref{table:mbv1_latency} and Table \ref{table:mbv2_latency} show that the prune networks discovered by MetaPruning achieve significantly higher accuracy than the uniform baselines with the same latency.
\setlength{\tabcolsep}{1pt}
\begin{table}
\begin{center}
\caption{This table compares the top-1 accuracy of MetaPruning method with the MobileNet V1~\cite{mobilenet_v1}, under the latency constraints. Reported latency is the run-time of the corresponding network on Titan Xp with a batch-size of 32}
\label{table:mbv1_latency}
\begin{tabular}{ccccccccc}
\noalign{\smallskip}
\hline
\multicolumn{3}{c}{\textbf{Uniform Baselines}} & \multicolumn{2}{c}{\textbf{MetaPruning}} \\
\hline
Ratio \ \ \ & Top1-Acc \ \ \ & Latency \ \ \ & Top1-Acc \ \ \ & Latency \ \ \ \\
\hline
\hline
1$\times$ & 70.6\% & 0.62ms & -- & --  \\
\hline
0.75$\times$ & 68.4\% & 0.48ms & \textbf{71.0\%} & 0.48ms\\
\hline
0.5$\times$ & 63.7\% & 0.31ms & \textbf{67.4\%} & 0.30ms\\
\hline
0.25$\times$ & 50.6\% & 0.17ms & \textbf{59.6\%} & 0.17ms\\
\hline
\vspace{-2em}
\end{tabular}
\end{center}
\end{table}
\setlength{\tabcolsep}{1.4pt}

\setlength{\tabcolsep}{1pt}
\begin{table}
\begin{center}
\caption{This table compares the top-1 accuracy of MetaPruning method with the MobileNet V2~\cite{mobilenet_v2}, under the latency constraints. We re-implement MobileNet V2 to obtain the results with 0.65 $\times$ and 0.35 $\times$ pruning ratio. This pruning ratio refers to uniformly prune the input and output channels of all the layers.}
\label{table:mbv2_latency}
\begin{tabular}{ccccccccc}
\noalign{\smallskip}
\hline
\multicolumn{3}{c}{\textbf{Uniform Baselines}} & \multicolumn{2}{c}{\textbf{MetaPruning}} \\
\hline
Ratio \ \ \ & Top1-Acc \ \ \ & Latency \ \ \ & Top1-Acc \ \ \ & Latency \ \ \ \\
\hline
\hline
1.4$\times$ & 74.7\% & 0.95ms & -- & --  \\
\hline
1$\times$ & 72.0\%  & 0.70ms & \textbf{73.2\%} & 0.67ms \\
\hline
0.65$\times$ & 67.2\% & 0.49ms & \textbf{71.7\%} & 0.47ms\\
\hline
0.35$\times$ & 54.6\% & 0.28ms & \textbf{64.5\%} & 0.29ms\\
\hline
\vspace{-3em}
\end{tabular}
\end{center}
\end{table}

\subsection{Pruned result visualization}
In channel pruning, people are curious about what is the best pruning heuristic and lots of human experts are working on manually designing the pruning policies. With the same curiosity, we wonder if any reasonable pruning schemes are learned by our MetaPruning method that contributes to its high accuracy. In visualizing the pruned network structures, we find that the MetaPruning did learn something interesting.

\begin{figure}[t]
\centering
\includegraphics[width=\linewidth]{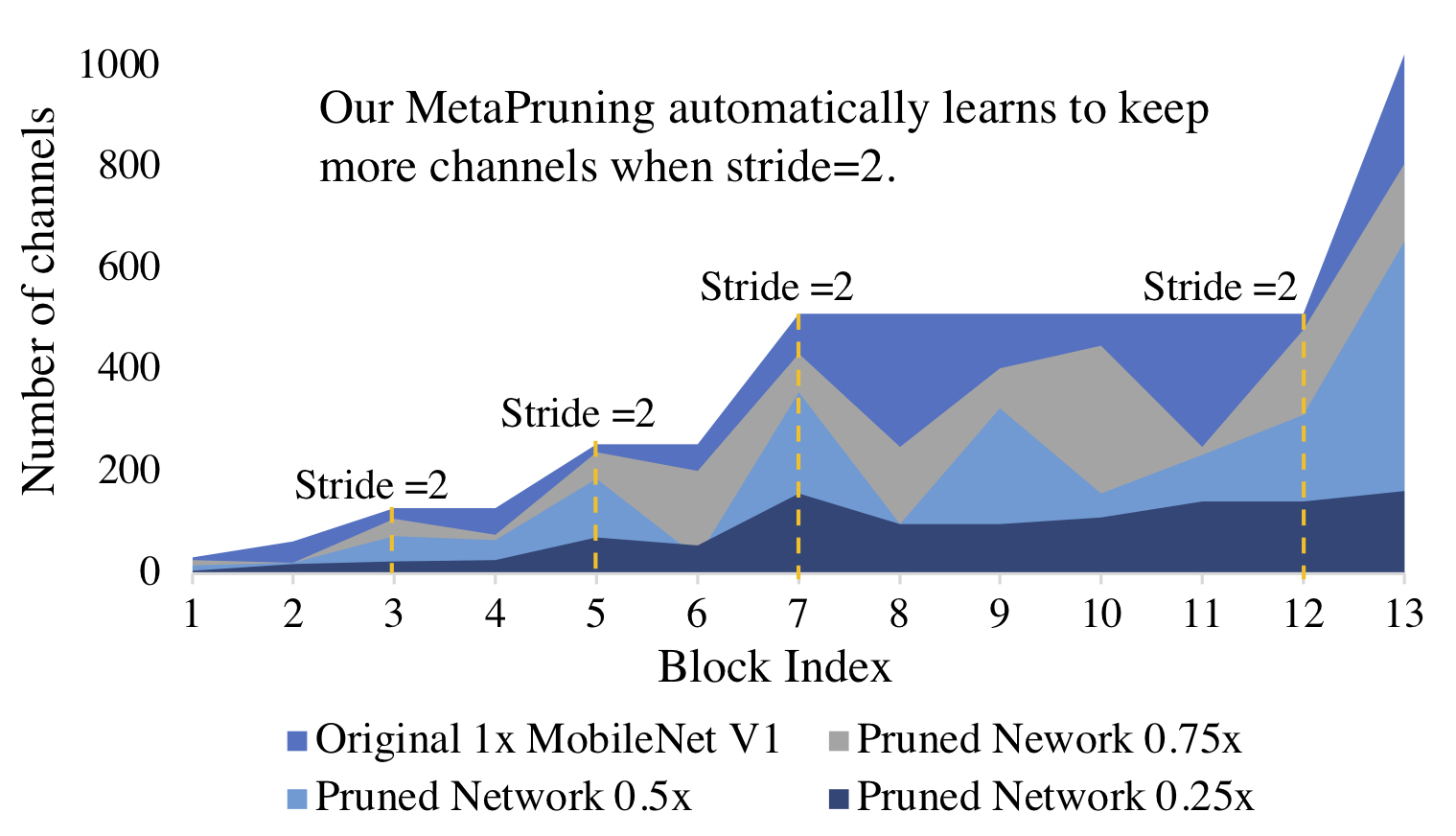}
\caption{This figure presents the number of output channels of each block of the pruned MobileNet v1. Each block contains a 3x3 depth-wise convolution followed by a 1x1 point-wise convolution, except the first block is composed by a 3x3 convolution only.}
\vspace{-1em}
\label{fig:vis_mbv1}
\end{figure}

\begin{figure}[t]
\centering
\includegraphics[width=\linewidth]{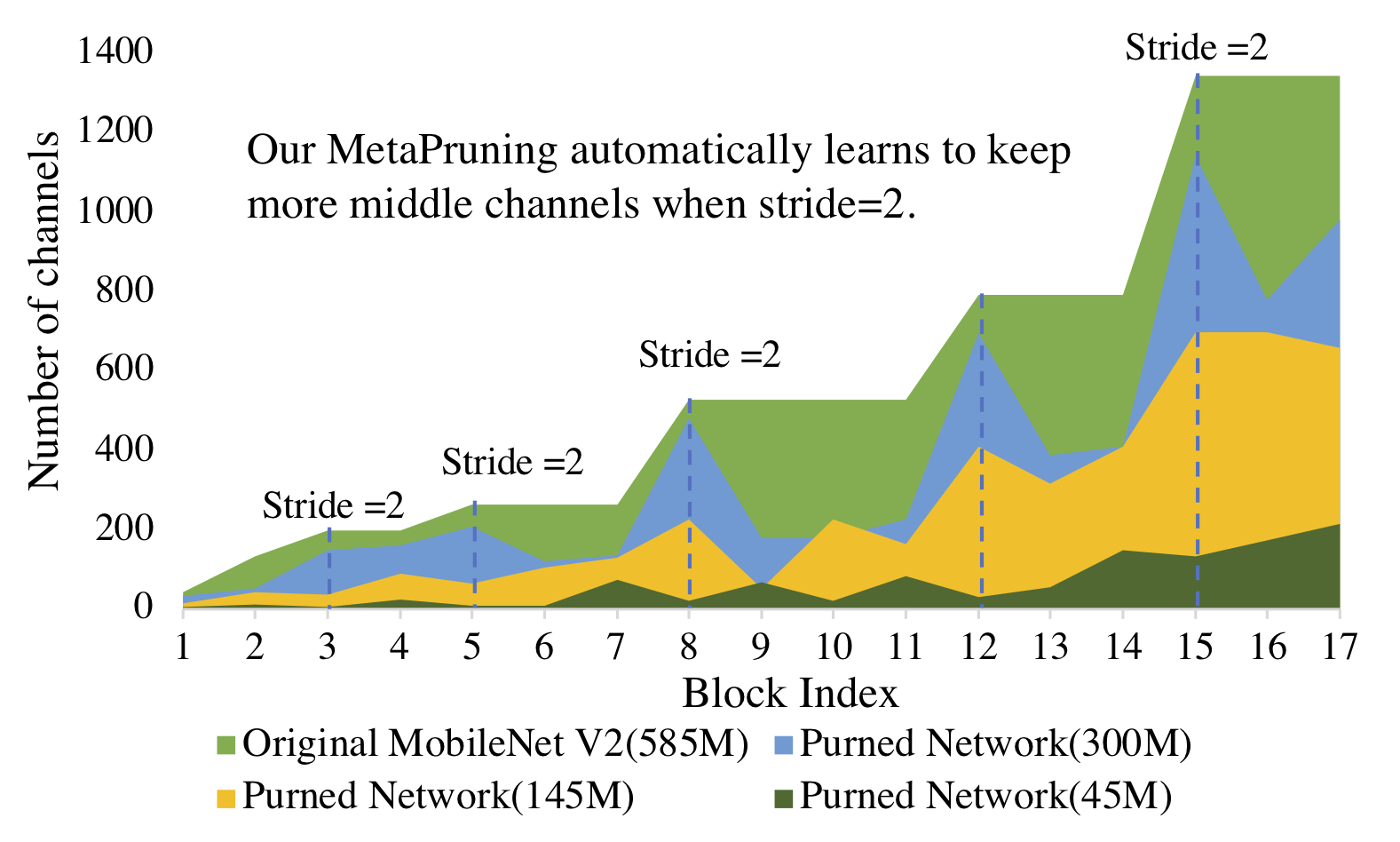}
\caption{A MobileNet V2 block is constructed by concatenating a 1x1 point-wise convolution, a 3x3 depth-wise convolution and a 1x1 point-wise convolution. This figure illustrates the number of middle channels of each block.}
\vspace{-1em}
\label{fig:vis_mbv2_mid}
\end{figure}

Figure \ref{fig:vis_mbv1} shows the pruned network structure of MobileNet V1. We observe significant peeks in the pruned network every time when there is a down sampling operation. When the down-sampling occurs with a stride 2 depth-wise convolution, the resolution degradation in the feature map size need to be compensated by using more channels to carry the same amount of information. Thus, MetaPruning automatically learns to keep more channels at the downsampling layers. The same phenomenon is also observed in MobileNet V2, shown in Figure \ref{fig:vis_mbv2_mid}. The middle channels will be pruned less when the corresponding block is in responsible for shrinking the feature map size.

Moreover, when we automatically prune the shortcut channels in MobileNet V2 with MetaPruning, we find that, despite the 145M pruned network contains only half of the FLOPs in the 300M pruned network, 145M network keeps similar number of channels in the last stages as the 300M network, and prunes more channels in the early stages. We suspect it is because the number of classifiers for the ImageNet dataset contains 1000 output nodes and thus more channels are needed at later stages to extract sufficient features. When the FLOPs being restrict to 45M, the network almost reaches the maximum pruning ratio and it has no choice but to prune the channels in the later stage, and the accuracy degradation from 145M network to 45M networks is much severer than that from 300M to 145M.

\begin{figure}[t]
\centering
\includegraphics[width=\linewidth]{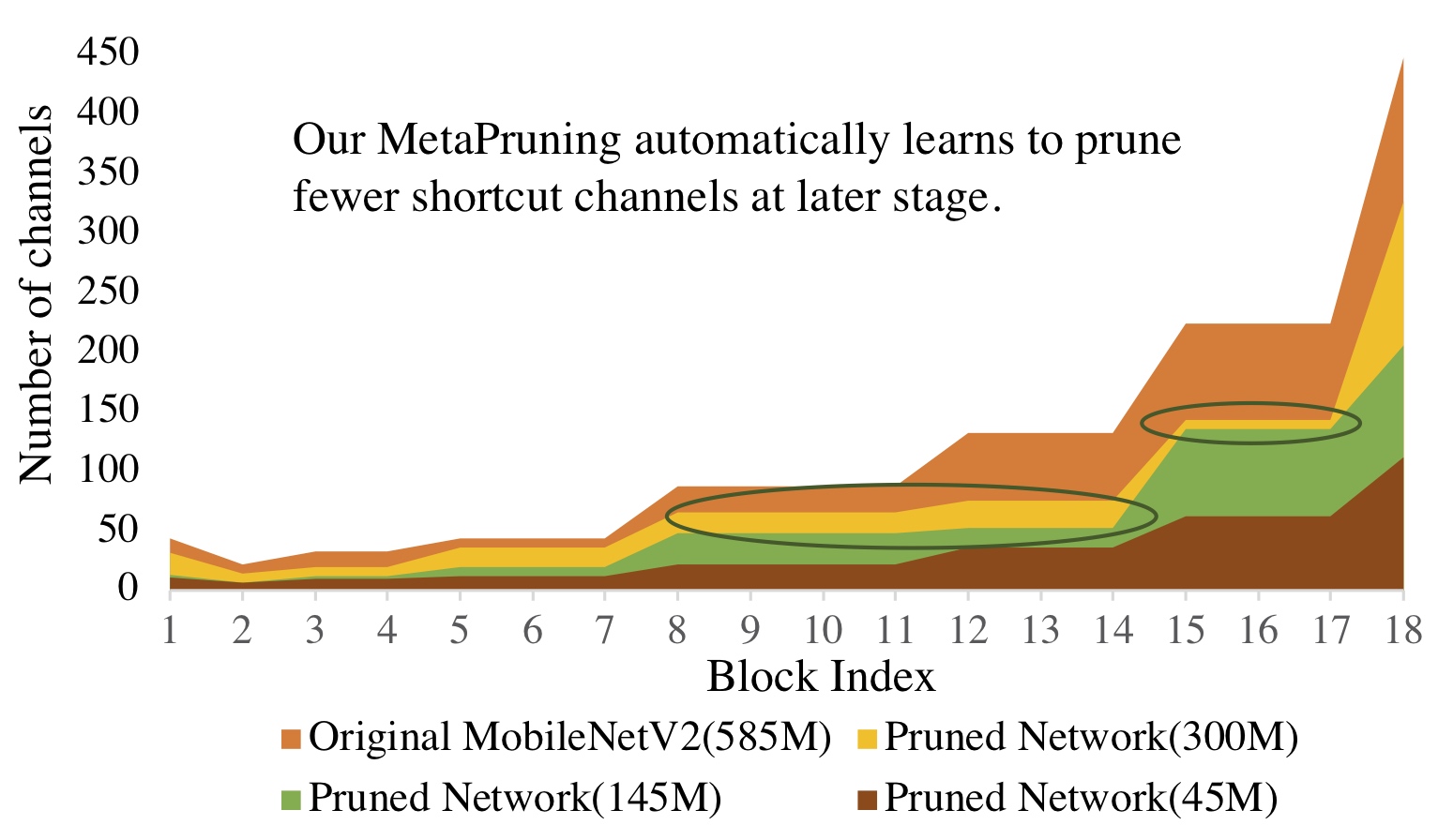}
\caption{In MobileNet V2, each stage starts with a bottleneck block with differed input and output channels and followed by several repeated bottleneck blocks. Those bottleneck blocks with the same input and output channels are connected with a shortcut. MetaPruning prunes the channels in the shortcut jointly with the middle channels. This figure illustrates the number of shortcut channel in each stage after being pruned by the MetaPruning.}
\vspace{-1em}
\label{fig:vis_mbv2_shortcut}
\end{figure}

\subsection{Ablation study}
In this section, we discuss about the effect of weight prediction in the MetaPruning method.

\begin{figure}[t]
\centering
\includegraphics[width=0.9\linewidth]{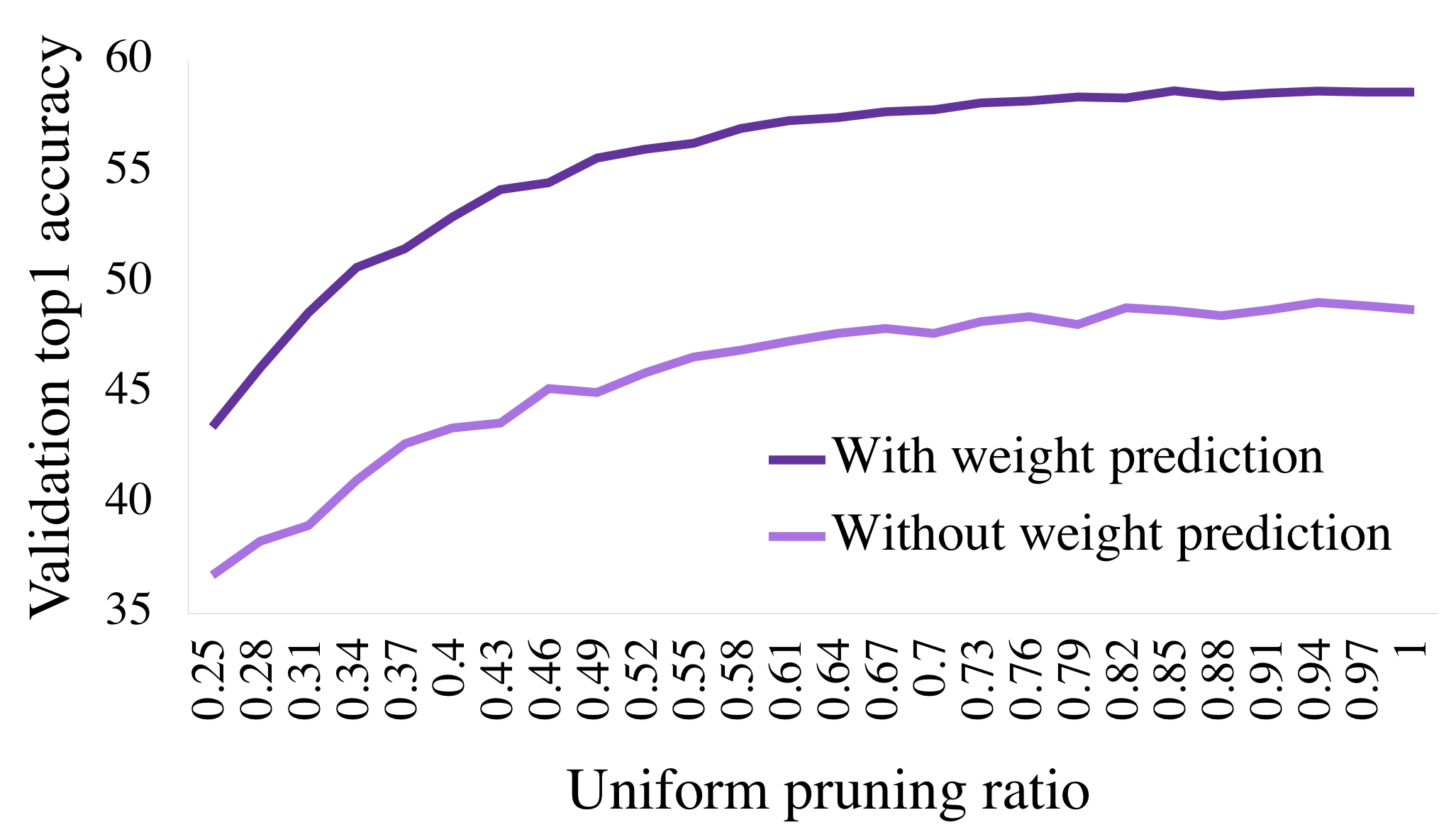}
\caption{We compare between the performance of PruningNet with weight prediction and that without weight prediction by inferring the accuracy of several uniformly pruned network of MobileNet V1\cite{mobilenet_v1}. PruningNet with weight prediction achieves much higher accuracy than that without weight prediction.}
\vspace{-1.5em}
\label{fig:meta_vs_nometa}
\end{figure}

We wondered about the consequence if we do not use the two fully-connected layers in the PruningNet for weight prediction but directly apply the proposed stochastic training and crop the same weight matrix for matching the input and output channels in the Pruned Network. We compare the performance between the PruningNet with and without weight prediction. We select the channel number with uniformly pruning each layer at a ratio ranging from [0.25, 1], and evaluate the accuracy with the weights generated by these two PruningNets. Figure \ref{fig:meta_vs_nometa} shows PruningNet without weight prediction achieves 10\% lower accuracy. We further use the PruningNet without weight prediction to search for the Pruned MobileNet V1 with less than 45M FLOPs. The obtained network achieves only 55.3\% top1 accuracy, 1.9\% lower than the pruned network obtained with weight prediction. It is intuitive. For example, the weight matrix for a input channel width of 64 may not be optimal when the total input channels are increased to 128 with 64 more channels added behind.
In that case, the weight prediction mechanism in meta learning is effective in de-correlating the weights for different pruned structures and thus achieves much higher accuracy for the PruningNet.
\vspace{-0.2em}

\section{Conclusion}
\vspace{-0.2em}
In this work, we have presented MetaPruning for channel pruning with following advantages: 1) it achieves much higher accuracy than the uniform pruning baselines as well as other state-of-the-art channel pruning methods, both traditional and AutoML-based; 2) it can flexibly optimize with respect to different constraints without introducing extra hyperparameters; 3) ResNet-like architecture can be effectively handled; 4) the whole pipeline is highly efficient.
\vspace{-0.2em}

\section{Acknowledgement}
\vspace{-0.2em}
The authors would like to acknowledge HKSAR RGC's funding support under grant GRF-16203918, National Key
R\&D Program of China (No. 2017YFA0700800) and Beijing Academy of Artificial Intelligence (BAAI).

{\small
\bibliographystyle{ieee_fullname}
\bibliography{egbib}

\begin{thebibliography}{10}\itemsep=-1pt

\bibitem{learning_number_of_neuron}
Jose~M Alvarez and Mathieu Salzmann.
\newblock Learning the number of neurons in deep networks.
\newblock In {\em Advances in Neural Information Processing Systems}, pages
  2270--2278, 2016.

\bibitem{structured_pruning}
Sajid Anwar, Kyuyeon Hwang, and Wonyong Sung.
\newblock Structured pruning of deep convolutional neural networks.
\newblock {\em ACM Journal on Emerging Technologies in Computing Systems
  (JETC)}, 13(3):32, 2017.

\bibitem{coarse_pruning}
Sajid Anwar and Wonyong Sung.
\newblock Compact deep convolutional neural networks with coarse pruning.
\newblock {\em arXiv preprint arXiv:1610.09639}, 2016.

\bibitem{Design_NN_using_RL}
Bowen Baker, Otkrist Gupta, Nikhil Naik, and Ramesh Raskar.
\newblock Designing neural network architectures using reinforcement learning.
\newblock {\em arXiv preprint arXiv:1611.02167}, 2016.

\bibitem{one_shot}
Gabriel Bender, Pieter-Jan Kindermans, Barret Zoph, Vijay Vasudevan, and Quoc
  Le.
\newblock Understanding and simplifying one-shot architecture search.
\newblock In {\em International Conference on Machine Learning}, pages
  549--558, 2018.

\bibitem{smash}
Andrew Brock, Theodore Lim, James~M Ritchie, and Nick Weston.
\newblock Smash: one-shot model architecture search through hypernetworks.
\newblock {\em arXiv preprint arXiv:1708.05344}, 2017.

\bibitem{proxylessnas}
Han Cai, Ligeng Zhu, and Song Han.
\newblock Proxylessnas: Direct neural architecture search on target task and
  hardware.
\newblock {\em arXiv preprint arXiv:1812.00332}, 2018.

\bibitem{chen2018constraint}
Changan Chen, Frederick Tung, Naveen Vedula, and Greg Mori.
\newblock Constraint-aware deep neural network compression.
\newblock In {\em Proceedings of the European Conference on Computer Vision
  (ECCV)}, pages 400--415, 2018.

\bibitem{chamnet}
Xiaoliang Dai, Peizhao Zhang, Bichen Wu, Hongxu Yin, Fei Sun, Yanghan Wang,
  Marat Dukhan, Yunqing Hu, Yiming Wu, Yangqing Jia, et~al.
\newblock Chamnet: Towards efficient network design through platform-aware
  model adaptation.
\newblock {\em arXiv preprint arXiv:1812.08934}, 2018.

\bibitem{imagenet}
Jia Deng, Wei Dong, Richard Socher, Li-Jia Li, Kai Li, and Li Fei-Fei.
\newblock Imagenet: A large-scale hierarchical image database.
\newblock 2009.

\bibitem{predicting_parameters}
Misha Denil, Babak Shakibi, Laurent Dinh, Nando De~Freitas, et~al.
\newblock Predicting parameters in deep learning.
\newblock In {\em Advances in neural information processing systems}, pages
  2148--2156, 2013.

\bibitem{dong2018dpp}
Jin-Dong Dong, An-Chieh Cheng, Da-Cheng Juan, Wei Wei, and Min Sun.
\newblock Dpp-net: Device-aware progressive search for pareto-optimal neural
  architectures.
\newblock In {\em Proceedings of the European Conference on Computer Vision
  (ECCV)}, pages 517--531, 2018.

\bibitem{zeroshot}
Mohamed Elhoseiny, Babak Saleh, and Ahmed Elgammal.
\newblock Write a classifier: Zero-shot learning using purely textual
  descriptions.
\newblock In {\em Proceedings of the IEEE International Conference on Computer
  Vision}, pages 2584--2591, 2013.

\bibitem{dynamic_surgery}
Yiwen Guo, Anbang Yao, and Yurong Chen.
\newblock Dynamic network surgery for efficient dnns.
\newblock In {\em Advances In Neural Information Processing Systems}, pages
  1379--1387, 2016.

\bibitem{hypernetworks}
David Ha, Andrew Dai, and Quoc~V Le.
\newblock Hypernetworks.
\newblock {\em arXiv preprint arXiv:1609.09106}, 2016.

\bibitem{deep_compression}
Song Han, Huizi Mao, and William~J Dally.
\newblock Deep compression: Compressing deep neural networks with pruning,
  trained quantization and huffman coding.
\newblock {\em arXiv preprint arXiv:1510.00149}, 2015.

\bibitem{learning_both_weights_and_connections}
Song Han, Jeff Pool, John Tran, and William Dally.
\newblock Learning both weights and connections for efficient neural network.
\newblock In {\em Advances in neural information processing systems}, pages
  1135--1143, 2015.

\bibitem{brain_surgeon}
Babak Hassibi, David~G Stork, and Gregory~J Wolff.
\newblock Optimal brain surgeon and general network pruning.
\newblock In {\em IEEE international conference on neural networks}, pages
  293--299. IEEE, 1993.

\bibitem{resnet}
Kaiming He, Xiangyu Zhang, Shaoqing Ren, and Jian Sun.
\newblock Deep residual learning for image recognition.
\newblock In {\em Proceedings of the IEEE conference on computer vision and
  pattern recognition}, pages 770--778, 2016.

\bibitem{he2018soft}
Yang He, Guoliang Kang, Xuanyi Dong, Yanwei Fu, and Yi Yang.
\newblock Soft filter pruning for accelerating deep convolutional neural
  networks.
\newblock {\em arXiv preprint arXiv:1808.06866}, 2018.

\bibitem{amc}
Yihui He, Ji Lin, Zhijian Liu, Hanrui Wang, Li-Jia Li, and Song Han.
\newblock Amc: Automl for model compression and acceleration on mobile devices.
\newblock In {\em Proceedings of the European Conference on Computer Vision
  (ECCV)}, pages 784--800, 2018.

\bibitem{channel_pruning}
Yihui He, Xiangyu Zhang, and Jian Sun.
\newblock Channel pruning for accelerating very deep neural networks.
\newblock In {\em Proceedings of the IEEE International Conference on Computer
  Vision}, pages 1389--1397, 2017.

\bibitem{loss-aware-quantization}
Lu Hou and James~T Kwok.
\newblock Loss-aware weight quantization of deep networks.
\newblock In {\em Proceedings of the International Conference on Learning
  Representations}, 2018.

\bibitem{mobilenet_v1}
Andrew~G Howard, Menglong Zhu, Bo Chen, Dmitry Kalenichenko, Weijun Wang,
  Tobias Weyand, Marco Andreetto, and Hartwig Adam.
\newblock Mobilenets: Efficient convolutional neural networks for mobile vision
  applications.
\newblock {\em arXiv preprint arXiv:1704.04861}, 2017.

\bibitem{network_trimming}
Hengyuan Hu, Rui Peng, Yu-Wing Tai, and Chi-Keung Tang.
\newblock Network trimming: A data-driven neuron pruning approach towards
  efficient deep architectures.
\newblock {\em arXiv preprint arXiv:1607.03250}, 2016.

\bibitem{learning_to_seg}
Ronghang Hu, Piotr Doll{\'a}r, Kaiming He, Trevor Darrell, and Ross Girshick.
\newblock Learning to segment every thing.
\newblock In {\em Proceedings of the IEEE Conference on Computer Vision and
  Pattern Recognition}, pages 4233--4241, 2018.

\bibitem{metasr}
Xuecai Hu, Haoyuan Mu, Xiangyu Zhang, Zilei Wang, Jian Sun, and Tieniu Tan.
\newblock Meta-sr: A magnification-arbitrary network for super-resolution.
\newblock {\em arXiv preprint arXiv:1903.00875}, 2019.

\bibitem{data-driven}
Zehao Huang and Naiyan Wang.
\newblock Data-driven sparse structure selection for deep neural networks.
\newblock In {\em Proceedings of the European Conference on Computer Vision
  (ECCV)}, pages 304--320, 2018.

\bibitem{squeezenet}
Forrest~N Iandola, Song Han, Matthew~W Moskewicz, Khalid Ashraf, William~J
  Dally, and Kurt Keutzer.
\newblock Squeezenet: Alexnet-level accuracy with 50x fewer parameters and< 0.5
  mb model size.
\newblock {\em arXiv preprint arXiv:1602.07360}, 2016.

\bibitem{brain_damage}
Yann LeCun, John~S Denker, and Sara~A Solla.
\newblock Optimal brain damage.
\newblock In {\em Advances in neural information processing systems}, pages
  598--605, 1990.

\bibitem{metalearning_survey}
Christiane Lemke, Marcin Budka, and Bogdan Gabrys.
\newblock Metalearning: a survey of trends and technologies.
\newblock {\em Artificial intelligence review}, 44(1):117--130, 2015.

\bibitem{pruning_filters}
Hao Li, Asim Kadav, Igor Durdanovic, Hanan Samet, and Hans~Peter Graf.
\newblock Pruning filters for efficient convnets.
\newblock {\em arXiv preprint arXiv:1608.08710}, 2016.

\bibitem{sparse_cnn}
Baoyuan Liu, Min Wang, Hassan Foroosh, Marshall Tappen, and Marianna Pensky.
\newblock Sparse convolutional neural networks.
\newblock In {\em Proceedings of the IEEE Conference on Computer Vision and
  Pattern Recognition}, pages 806--814, 2015.

\bibitem{darts}
Hanxiao Liu, Karen Simonyan, and Yiming Yang.
\newblock Darts: Differentiable architecture search.
\newblock {\em arXiv preprint arXiv:1806.09055}, 2018.

\bibitem{network_slimming}
Zhuang Liu, Jianguo Li, Zhiqiang Shen, Gao Huang, Shoumeng Yan, and Changshui
  Zhang.
\newblock Learning efficient convolutional networks through network slimming.
\newblock In {\em Proceedings of the IEEE International Conference on Computer
  Vision}, pages 2736--2744, 2017.

\bibitem{rethink_pruning}
Zhuang Liu, Mingjie Sun, Tinghui Zhou, Gao Huang, and Trevor Darrell.
\newblock Rethinking the value of network pruning.
\newblock {\em arXiv preprint arXiv:1810.05270}, 2018.

\bibitem{birealnet}
Zechun Liu, Baoyuan Wu, Wenhan Luo, Xin Yang, Wei Liu, and Kwang-Ting Cheng.
\newblock Bi-real net: Enhancing the performance of 1-bit cnns with improved
  representational capability and advanced training algorithm.
\newblock In {\em Proceedings of the European Conference on Computer Vision
  (ECCV)}, pages 722--737, 2018.

\bibitem{thinet}
Jian-Hao Luo, Jianxin Wu, and Weiyao Lin.
\newblock Thinet: A filter level pruning method for deep neural network
  compression.
\newblock In {\em Proceedings of the IEEE international conference on computer
  vision}, pages 5058--5066, 2017.

\bibitem{shufflenet_v2}
Ningning Ma, Xiangyu Zhang, Hai-Tao Zheng, and Jian Sun.
\newblock Shufflenet v2: Practical guidelines for efficient cnn architecture
  design.
\newblock In {\em Proceedings of the European Conference on Computer Vision
  (ECCV)}, pages 116--131, 2018.

\bibitem{diversity_networks}
Zelda Mariet and Suvrit Sra.
\newblock Diversity networks.
\newblock {\em Proceedings of ICLR}, 2016.

\bibitem{pruning_CNN}
Pavlo Molchanov, Stephen Tyree, Tero Karras, Timo Aila, and Jan Kautz.
\newblock Pruning convolutional neural networks for resource efficient transfer
  learning.
\newblock {\em arXiv preprint arXiv:1611.06440}, 3, 2016.

\bibitem{ENAS}
Hieu Pham, Melody~Y Guan, Barret Zoph, Quoc~V Le, and Jeff Dean.
\newblock Efficient neural architecture search via parameter sharing.
\newblock {\em arXiv preprint arXiv:1802.03268}, 2018.

\bibitem{xnornet}
Mohammad Rastegari, Vicente Ordonez, Joseph Redmon, and Ali Farhadi.
\newblock Xnor-net: Imagenet classification using binary convolutional neural
  networks.
\newblock In {\em European Conference on Computer Vision}, pages 525--542.
  Springer, 2016.

\bibitem{fewshot}
Sachin Ravi and Hugo Larochelle.
\newblock Optimization as a model for few-shot learning.
\newblock 2016.

\bibitem{large_scale_evolution}
Esteban Real, Sherry Moore, Andrew Selle, Saurabh Saxena, Yutaka~Leon Suematsu,
  Jie Tan, Quoc~V Le, and Alexey Kurakin.
\newblock Large-scale evolution of image classifiers.
\newblock In {\em Proceedings of the 34th International Conference on Machine
  Learning-Volume 70}, pages 2902--2911. JMLR. org, 2017.

\bibitem{mobilenet_v2}
Mark Sandler, Andrew Howard, Menglong Zhu, Andrey Zhmoginov, and Liang-Chieh
  Chen.
\newblock Mobilenetv2: Inverted residuals and linear bottlenecks.
\newblock In {\em Proceedings of the IEEE Conference on Computer Vision and
  Pattern Recognition}, pages 4510--4520, 2018.

\bibitem{efficient_dnn_survey}
Vivienne Sze, Yu-Hsin Chen, Tien-Ju Yang, and Joel~S Emer.
\newblock Efficient processing of deep neural networks: A tutorial and survey.
\newblock {\em Proceedings of the IEEE}, 105(12):2295--2329, 2017.

\bibitem{learning_to_learn}
Yu-Xiong Wang and Martial Hebert.
\newblock Learning to learn: Model regression networks for easy small sample
  learning.
\newblock In {\em European Conference on Computer Vision}, pages 616--634.
  Springer, 2016.

\bibitem{fbnet}
Bichen Wu, Xiaoliang Dai, Peizhao Zhang, Yanghan Wang, Fei Sun, Yiming Wu,
  Yuandong Tian, Peter Vajda, Yangqing Jia, and Kurt Keutzer.
\newblock Fbnet: Hardware-aware efficient convnet design via differentiable
  neural architecture search.
\newblock {\em arXiv preprint arXiv:1812.03443}, 2018.

\bibitem{genetic_CNN}
Lingxi Xie and Alan Yuille.
\newblock Genetic cnn.
\newblock In {\em Proceedings of the IEEE International Conference on Computer
  Vision}, pages 1379--1388, 2017.

\bibitem{metaanchor}
Tong Yang, Xiangyu Zhang, Zeming Li, Wenqiang Zhang, and Jian Sun.
\newblock Metaanchor: Learning to detect objects with customized anchors.
\newblock In {\em Advances in Neural Information Processing Systems}, pages
  318--328, 2018.

\bibitem{netadapt}
Tien-Ju Yang, Andrew Howard, Bo Chen, Xiao Zhang, Alec Go, Mark Sandler,
  Vivienne Sze, and Hartwig Adam.
\newblock Netadapt: Platform-aware neural network adaptation for mobile
  applications.
\newblock In {\em Proceedings of the European Conference on Computer Vision
  (ECCV)}, pages 285--300, 2018.

\bibitem{slimmable}
Jiahui Yu, Linjie Yang, Ning Xu, Jianchao Yang, and Thomas Huang.
\newblock Slimmable neural networks.
\newblock {\em arXiv preprint arXiv:1812.08928}, 2018.

\bibitem{lq-net}
Dongqing Zhang, Jiaolong Yang, Dongqiangzi Ye, and Gang Hua.
\newblock Lq-nets: Learned quantization for highly accurate and compact deep
  neural networks.
\newblock In {\em Proceedings of the European Conference on Computer Vision
  (ECCV)}, pages 365--382, 2018.

\bibitem{shufflenet_v1}
Xiangyu Zhang, Xinyu Zhou, Mengxiao Lin, and Jian Sun.
\newblock Shufflenet: An extremely efficient convolutional neural network for
  mobile devices.
\newblock In {\em Proceedings of the IEEE Conference on Computer Vision and
  Pattern Recognition}, pages 6848--6856, 2018.

\bibitem{elq}
Aojun Zhou, Anbang Yao, Kuan Wang, and Yurong Chen.
\newblock Explicit loss-error-aware quantization for low-bit deep neural
  networks.
\newblock In {\em Proceedings of the IEEE Conference on Computer Vision and
  Pattern Recognition}, pages 9426--9435, 2018.

\bibitem{zhuang2019structured}
Bohan Zhuang, Chunhua Shen, Mingkui Tan, Lingqiao Liu, and Ian Reid.
\newblock Structured binary neural networks for accurate image classification
  and semantic segmentation.
\newblock In {\em Proceedings of the IEEE Conference on Computer Vision and
  Pattern Recognition}, pages 413--422, 2019.

\bibitem{NAS_with_RL}
Barret Zoph and Quoc~V Le.
\newblock Neural architecture search with reinforcement learning.
\newblock {\em arXiv preprint arXiv:1611.01578}, 2016.

\end{thebibliography}
}

\end{document}